\title{A Batchwise Monotone Algorithm \\for Dictionary Learning}
\author{
        Huan Wang \\
                	Yahoo!Labs, New York\\
            \and
        John Wright \\
        Computer Science Department, Columbia University,\\
        \and 
        Daniel Spielman\\
        Computer Science Department, Yale University
}
\date{\today}

\documentclass{article} % Include author names

\usepackage{times,amsmath,amsfonts,amsthm,amssymb, verbatim, algorithm, algorithmic,amsthm,color,graphicx,epsfig, subfigure,verbatim,url,fancybox,url}

\newenvironment{definition}[1][Definition]{\begin{trivlist}
\item[\hskip \labelsep {\bfseries #1}]}{\end{trivlist}}

\newtheorem{theorem}{Theorem}[section]
\newtheorem{lemma}[theorem]{Lemma}

\begin{document}

\maketitle

\begin{abstract}
We propose a batchwise monotone algorithm for dictionary
learning. Unlike the state-of-the-art dictionary learning algorithms
which impose sparsity constraints on a sample-by-sample basis, we
instead treat the samples as a batch, and impose the sparsity
constraint on the whole. The benefit of batchwise optimization is that
the non-zeros can be better allocated across the samples, leading to a better approximation of the whole. To accomplish this, we propose procedures to switch non-zeros in both rows
and columns in the support of the coefficient matrix to reduce the
reconstruction error. We prove in the proposed support switching
procedure the objective of the algorithm, i.e., the reconstruction
error, decreases monotonically and converges. Furthermore, we introduce
a block orthogonal matching pursuit algorithm that also operates
on sample batches to provide a warm start. Experiments on both natural
image patches and UCI data sets show that the proposed algorithm
produces a better approximation with the same sparsity levels compared
to the state-of-the-art algorithms.

\end{abstract}

\section{Introduction} 

A number of algorithms have recently been developed that
automatically design representations through a process called dictionary learning. The hope is that learning algorithms can exploit structure in specific classes of signals, enabling better performance in applications. Dictionary learning algorithms have already been used successfully in a number of image processing problems, such as image
compression \cite{Aharon05}\cite{Skretting11}, inpainting
\cite{Mairal09}\cite{Zhou09}, image denoising
\cite{Aharon05}\cite{Li12}\cite{Elad06}\cite{Elad08}, super-resolution
\cite{Yang10}\cite{Zhou12}, digit recognition, and texture
classification \cite{Mairal08}.

Popular dictionary learning algorithms can be roughly divided into two
categories: hard constraint-based \cite{Aharon05} \cite{Engan1999-ICASSP}, and soft
sparsity-penalty-based \cite{Mairal09}\cite{Olshausen96}\cite{Lee06}.
These algorithms search for a dictionary of vectors (called atoms)
so that it is possible to represent each sample signal
as a linear combination of a small number of the atoms.
Often dictionaries with more atoms than the dimension, called over-complete
  dictionaries, are used.

Since we would like to use only a few atoms in the representation of each sample, a sparsity constraint is imposed on the coefficients in the representations. Both K-SVD \cite{Aharon05} and the online dictionary learning \cite{Mairal09} algorithm impose sparsity constraints, either hard or soft, on the representation of individual samples. However, it may not be optimal to assume a similar sparsity level for each sample. In fact, some samples could be easy to represent and some may require more atoms in their representations. The recent dictionary learning algorithm of \cite{Spielman12} searches for dictionaries that have a sparsity constraint on the number of times each atom is used. Thus, some signals can be represented using more atoms than others. Their algorithm inspires us to focus on how individual atoms are used rather than how individual signals are represented.

In this paper we present a monotone algorithm for dictionary learning. Similar to \cite{Spielman12}, the algorithm we propose in this work also acts on the rows of the coefficient matrix, but can empirically produce good approximations even in the more challenging (and realistic) conditions. In contrast to the traditional sample-based sparsity constraint, we impose the sparsity constraint in a batchwise fashion. That is, we switch the positions of the non-zeros in the coefficients within batches of samples among different columns and rows in the coefficient matrix, and at the same time keep the total number of non-zeros fixed. As a result, the number of non-zeros, constrained within a batch of samples, is allowed to vary in either columns or rows . We show that all the non-zero position switching operations only reduce reconstruction error, leading to a convergent objective function. For initialization, we introduce a simple iterative dictionary update procedure that operates on a batch of samples to give an approximate guess of the dictionary. In each iteration, first the non-zero patterns are derived using a block orthogonal matching pursuit and then the dictionary is updated using least squares. 

There are two main advantages of our proposed algorithm:
\begin{enumerate}
\item Since the non-zero positions are optimized in a batchwise fashion, we are able to achieve a smaller reconstruction error, or better approximation, compared to the traditional sample-by-sample constraint with the same level of sparsity.
\item The reconstruction error is guaranteed to decrease monotonically and converge.
\end{enumerate}

\section{Notation}\label{sec:note}

In the dictionary learning problem, 
  one is given a matrix that contains the sample signals in its columns, 
   $Y=[y_1,y_2,\dots,y_p]\in \mathbb{R}^{m\times p}$,
  along with a target number of atoms, $n$.
The goal is to find a dictionary of atoms $A\in \mathbb{R}^{m\times n}$
  and a sparse coefficient matrix $X\in \mathbb{R}^{n\times p}$
  so that $Y\approx AX$.

Throughout this paper, $m$ is the dimension of
  each sample and $p$ is the number of samples.
We use $a_i$
and $x_i$ to denote the $i$th column of $A$ and $X$ respectively, and
$x^i$ to denote the $i$th row of $X$. We use $\Omega^i$ to denote the
support of $x^i$, $\Omega_i$ for the support of $x_i$,
$k^i=|\Omega^i|$, and $k_i=|\Omega_i|$. $\otimes$ is the Kronecker product, $I_p$ is a $p$-by-$p$ identity matrix, and $y=\mathrm{vec}(Y)$, where $\mathrm{vec}(\cdot)$ concatenates the columns of $Y$ to form a vector.

For a set 
$\Omega \subseteq \{1, \dotsc , \} $, we let
$P_\Omega$ to denote the projection matrix
%\begin{align}
 $P_{\Omega} = [e_{\omega_1},e_{\omega_2},\dots,e_{\omega_{|\Omega|}}]$,
 %\end{align}
where $e_i$ is the elementary unit vector in coordinate $i$. 
We use $Y_\Omega$ to denote $YP_\Omega$. $A/a_i$ means the sub-matrix constructed by removing the $i$th column of $A$, and $X/x^i$ is the sub-matrix constructed by removing the $i$th row of $X$. When describing iterative algorithms, we use $x^{(+)}$ to denote the updated value of $x$. 

\section{Columnwise Sparsity Constraints}\label{sec:prev}

State-of-the-art dictionary learning algorithms treat the input sample matrix $Y$ in a sample-by-sample way. That is, the sparseness constraint is imposed on the coefficient of each sample independently using the current dictionary, and then the dictionary and coefficient matrix are updated accordingly. Among popular dictionary learning algorithms, two representative ones are the hard-constraint-based $K$-SVD \cite{Aharon05} and the soft-penalty-based dictionary learning algorithms such as the online learning algorithm \cite{Mairal09} and the efficient sparse coding algorithms \cite{Lee06}.

The $K$-SVD algorithm aims to iteratively minimize the objective 
\begin{align}
\min_{A,X} \|Y-AX\|_f^2~~\mathrm{s.t.}~~\forall i, \|x_i\|_0\leq k,
\end{align}
where $A$ is the dictionary and $X$ is the sparse coefficient matrix.

Empirically the $K$-SVD algorithms often works well, but its objective value is not guaranteed to decrease monotonically because the support of $X$ is changed using the greedy pursuit algorithm one column at a time.

%\subsection{Online Dictionary Learning}

The algorithms in \cite{Mairal09} \cite{Lee06} replace the hard $\ell_0$ penalty with an $\ell_1$ penalty, giving
\begin{align}
\min_{A,X} \|Y-AX\|_f^2+\lambda\sum_i \|x_i\|_1.
\end{align}
This optimization problem is not convex. However, it is convex in $X$ if $A$ is fixed, or in $A$ if $X$ is fixed. As in the method of optimal directions (MOD), the optimization is done via alternating directions.

The advantage of column-wise sparsity constraints is that it leads naturally to fast online algorithms. Whenever a new sample comes one simply adds a sparsity constraint on the incoming column of $X$. The downside is the sample-by-sample sparsity treatment lacks a ``global'' view of the sparsity pattern. For example, there is no reason we should require each sample to be represented  by exactly $k$ atoms in the dictionary, or impose a sparsity penalty with the same $\lambda$. Some samples, being ``harder" to approximate, require more atoms, and it could be a waste to use too many atoms to represent the ``easy'' samples.

\section{Batchwise Support Switching Procedures}\label{sec:suppSwitch}

Unlike $K$-SVD and online learning, which constrain the sparsity of the coefficients in a column-by-column fashion, we argue that it may be possible to obtain better sparse approximations of the input as a whole if we allow the column sparsity to vary. 
We seek the best possible reconstruction, subject to a constraint on the {\em total} number of nonzeros across the batch of samples:
\begin{align}
\min_{A,X} \|Y-AX\|_f^2 ~~\mathrm{s.t.}~~ \|X\|_0\leq K,
\end{align}
The advantage is that some non-zero positions with less impact on the objective can be replaced using crucial ones. As a result a more accurate decomposition is produced with different column sparsities across samples.

We introduce a heuristic for attacking this problem, which computes an initial sparsifying dictionary using alternating directions, and then refines it using sequence of support and amplitude adjustments. The initial approximation makes use of a {\em batchwise} orthogonal matching pursuit, which aims at minimizing the $\ell_0$ norm of $X$ as a whole.

The support switching procedure updates the non-zero positions, i.e., the sparsity patterns, in the coefficient matrix $X$ in two ways: inner-row switching and inter-row switching. In the inner-row switching, the total number of non-zeros in each row is fixed, and the non-zero positions are adjusted within the same row; and in the inter-row switching, the total number of non-zeros in pairs of rows is fixed, and the non-zeros are changed between two rows. Finally, we introduce an iterative procedure to adjust the amplitude of the coefficient and dictionary, with the sparsity pattern fixed. The whole algorithm is described in Algorithm \ref{alg:monosvd}.  
We prove that in the procedure of sparsity pattern switching and the amplitude adjustment, the objective decreases monotonically and converges. 

\begin{algorithm}[tb]
   \caption{Inner Row Support Switching}
   \label{proc:innerRow}
\begin{algorithmic}
   \STATE {\bfseries Input:} $\tilde{Y}\in \mathbb{R}^{m\times p}$, $x\in \mathbb{R}^{1\times p}$, $a\in \mathbb{R}^{m\times 1}$, and $N$. 
   \STATE {\bfseries Output:} $a^{(+)}\in \mathbb{R}^{m\times 1}$, and $x^{(+)}\in \mathbb{R}^{1\times p}$. 
   \STATE Denote the support  of $x$ as $\Omega$, and $k=|\Omega|$ .
%    \REPEAT $n$ times
   \FOR{$i=1$ {\bfseries to} $N$}
    \STATE SVD: $\tilde{Y}_{\Omega}=\sigma_1  ax_{\Omega}+\sum_{1<j\leq m} \sigma_j u_jv_j^T$, s.t., $\sigma_1\geq \sigma_2\dots\geq \sigma_m$, and $\|a\|_2=1$.
    \STATE Use the indices of the $k$ largest  $\{|a^T\tilde{y}_i|$\} as $\Omega$.
    \STATE $x_{\Omega}=a^{T}\tilde{Y}_{\Omega^{}}/\|a\|_2^2$,  $x_{{\Omega}^\mathsf{c}}=0$.
   \ENDFOR
   \STATE $a^{(+)}=a$, $x^{(+)}=x$.
%   \UNTIL{$\|\tilde{Y}-ax\|_2^2-\|Y-A^{(+)}X^{(+)}\|_2^2\leq \epsilon$}
\end{algorithmic}
\end{algorithm}

\subsection{Inner-Row Support Switching}

Suppose we are given the number of non-zeros $k^i$ in each row of $X$. The problem becomes
\begin{align}\label{alg:obj}
\min_{A,X} \|Y-AX\|_f^2~~~\mathrm{s.t.}~~~\|x^i\|_0=k^i, \;\; i = 1, \dots, p.
\end{align}
Note here $x^i$ is the $i$-th row of $X$ compared to the $i$-th column in the $K$-SVD objective. Globally optimizing this objective is challenging, due to the nonconvexity of the constraint set and the objective. Similar to $K$-SVD, we can obtain a simpler subproblem by only considering one row at a time, giving 
\begin{equation}
\min_{a_i, x^i} \; \bigl\| Y - \sum_{j\neq i} a_j x^j -a_ix^i\bigr\|_f^2 ~~~\mathrm{s.t.}~~~\|x^i \|_0 \le k^i.
\end{equation}
Setting $\tilde{Y} = Y - \sum_{j \ne i} a_j x^j$, the problem becomes one of finding a rank-one approximation to $\tilde{Y}$, with at most $k^i$ nonzero columns. 

We attack this problem using alternating directions. Assuming the support $\Omega^i$ is known and fixed, a best rank one approximation $a_i x^i$ can be fit using the SVD of $\tilde{Y}_{\Omega^i}$. If, on the other hand $a_i$ is fixed, an optimal support $\Omega^i$ can be derived by simply ranking the absolute values of the projected samples, i.e., $|a_i^Ty_i|$. Once the support $\Omega^i$ and the atom $a_i$ are known, $x^i$ can be calculated in closed-form. The resulting algorithm is listed in Algorithm \ref{proc:innerRow}.

\subsection{Inter-Row Support Switching}\label{sec:inter-adj}
\begin{algorithm}[tb]
   \caption{Inter Row Support Switching}
   \label{proc:interRow}
\begin{algorithmic}
   \STATE {\bfseries Input:} $\tilde{Y}\in \mathbb{R}^{m\times p}$, $x^i, x^j\in \mathbb{R}^{1\times p}$, and $a_i,a_j\in \mathbb{R}^{m\times 1}$. 
   \STATE {\bfseries Output:} $x^{i(+)}, x^{j(+)} \in \mathbb{R}^{1\times p}$.
   \STATE 1. Denote the support  of $x^i$ and $x^j$ as $\Omega^i$, and $\Omega^j$ respectively, and $\Omega=(\Omega^i\cup\Omega^j)\setminus(\Omega^i\cap \Omega^j)$. Denote $\tilde{\Omega}={(\Omega_i\cap\Omega_j)}^\mathsf{c}$.%the support of $X^{(+)}_{{(\Omega_i\cap\Omega_j)}^\mathsf{c}}=[x^i;x^j]_{{(\Omega_i\cap\Omega_j)}^\mathsf{c}}$ as $\Omega_e$.
%    \REPEAT $n$ times
%   \FOR{$i=1$ {\bfseries to} $N$}
    \STATE 2. Form the matrix $M=[a_i,a_j]^T\tilde{Y}_{\tilde{\Omega}}$. 
    \STATE 3. Pick up the larger entry of the two rows in each column in $|M|$ as candidates, and use the positions of the largest $|\Omega|$ candidates as $\Omega^{(+)}$. 
    \STATE 4. Set $X_{\Omega^{(+)}}=M_{\Omega^{(+)}}$, $X_{\tilde{\Omega}\setminus\Omega^{(+)}}=0$, and $X_{\Omega_i\cap\Omega_j}=[x^i;x^j]_{\Omega_i\cap\Omega_j}$. 
    \STATE 5.  Return $x^{i(+)}=e_1^TX$ and $x^{j(+)}=e_2^TX$.
\end{algorithmic}
\end{algorithm}

In this section, we introduce a procedure to adjust the non-zeros between two rows of $X$, such that the reconstruction error in (\ref{alg:obj}) decreases and at the same the total number of non-zeros in the two rows stays the same. First, we define the unique columns in $X_s=[x^i;x^j]$ to be the symmetric difference of the supports of $x^i$ and $x^j$:
\begin{definition} Suppose the support of $x^i$ and $x^j$ are $\Omega^i$ and $\Omega^j$ respectively, then the index set of the unique columns in $\begin{bmatrix}x^i\\x^j\end{bmatrix}$ are
$\Omega=(\Omega^i\cup\Omega^j)\setminus(\Omega^i\cap \Omega^j).$
\end{definition}

If we fix the remaining columns, the residual is 
$\tilde{Y}=Y-[A\setminus \{a_i,a_j\}][X\setminus \{x^i,x^j\}].$  
Again, if we fix the dictionary atoms $a_i$ and $a_j$, if $\|a_i\|=\|a_j\|=1$, the optimal support for the $|\Omega|$ unique columns can be derived by ranking the absolute values of the projected sample $M=[a_i,a_j]^T\tilde{Y}_{{(\Omega_i\cap\Omega_j)}^\mathsf{c}}$ with the constraint that we can only pick up one non-zero in each column of $M$. The procedure is described in Algorithm \ref{proc:interRow}.

The inter-row support switching reduces the objective in (\ref{alg:obj}) by fixing the dictionary and comparing the importance of non-zero positions by ranking the projected absolute values of the residuals $\tilde{Y}$. If we run the procedure for all pairs of rows in $X$, the total number of non-zeros in $X$ stays the same but their distribution is optimized batchwisely. The procedure is based on fixed dictionary $A$, and the optimization is only carried out on rows of $X$. 
Switching supports between all pairs of rows can be expensive when the number of rows $n$ in $X$ grows large. In application, we use two ways to reduce the computation cost:
\begin{enumerate}
\item Instead of going over all pairs of rows in $X$, we only go through a randomly sampled subset of the $(^n_2)$ pairs.
\item The inter-row support switching is only carried out when the objective decreases very slowly in the inner-row support switching.
\end{enumerate}

The inner-row and inter-row support switching interchange the positions of the non-zeros in the coefficient matrix within a batch of samples. In the following section we will introduce a procedure to further reduce the objective by changing the amplitude of the entries in both $A$ and $X$ given the support $\Omega$ of $X$. %operating on both $A$ and $X$. 

\section{Alternating Amplitude Adjustment}\label{sec:globalAdj}

\vskip -0.1in\begin{algorithm}%[tb]
   \caption{Alternating Amplitude Adjustment}
   \label{proc:glAdj}
\begin{algorithmic}
   \STATE {\bfseries Input:} $Y\in \mathbb{R}^{m\times p}$, $X \in \mathbb{R}^{n\times p}$, $A\in \mathbb{R}^{m\times n}$, and $N$. 
   \STATE {\bfseries Output:} $X^{(+)}\in \mathbb{R}^{n\times p}$, and $A^{(+)} \in \mathbb{R}^{m\times n}$.
   \STATE Denote the support of $X$ as $\Omega$, and the support of $x_i$ as $\Omega_i$.
   \FOR{$i=1$ {\bfseries to} $N$}
    \STATE $A=YX^T(XX^T)^{-1}$. 
    \FOR{$j=1$ {\bfseries to} $p$}
  	  \STATE $x_j(\Omega_j)=(A_{\Omega_j}^TA_{\Omega_j})^{-1}A_{\Omega_j}^Ty_j$
  	  \STATE $x_j(\Omega_j^\mathsf{c})=0$  	  
	\ENDFOR
    \ENDFOR
 \STATE $X^{(+)}=X$, $A^{(+)}=A$.
\end{algorithmic}
\end{algorithm}

In this section, we propose an alternating optimization algorithm for the following problem:
\begin{align}
\min_{A,X} ~~\|Y-AX\|_f^2, ~~~~~\mathrm{s.t.}~~~ X(\Omega^\mathsf{c})=0
\end{align}
If $X$ is known and fixed, the optimal $A$ can be computed via least squares:
%\begin{align}
$A=YX^T(XX^T)^{-1}$.
%\end{align}

On the other hand, given $A$ and $\Omega$, the objective above decomposes into a sum of samplewise reconstruction errors:
%\begin{align}
 $\min_{X_\Omega}\|Y-AX\|_f^2=\min_{x_j(\Omega_j)}\sum_j\|y_j-A_{\Omega_j}x_j(\Omega_j)\|_2^2$,
%\end{align}
which amounts to solving least squares for each column of $X$ sample-by-sample, that is:
%\begin{align}
$x_j(\Omega_j)=(A_{\Omega_j}^TA_{\Omega_j})^{-1}A_{\Omega_j}^T y_j, ~~\mathrm{and}~~x_j(\Omega_j^\mathsf{c})=0$.
%\end{align}
The detailed procedure is shown in Algorithm \ref{proc:glAdj}. It is not hard to see that in each iteration the objective does not increase.

\section{Proof of Monotonicity}\label{sec:proof}%{The Monotone Dictionary Learning Algorithm}

The full procedure is described in Algorithm \ref{alg:monosvd}. \footnote{Ways to reduce the computation cost is discussed in section \ref{sec:inter-adj}}. We will show in this section that all the procedures introduced in section (\ref{sec:suppSwitch}) and section (\ref{sec:globalAdj}) only decrease the objective value, while keeping the total number of nonzero coefficients unchanged. 

\begin{algorithm}[tb]
   \caption{BatchSVD}
   \label{alg:monosvd}
\begin{algorithmic}
   \STATE {\bfseries Input:} $Y\in \mathbb{R}^{m\times p}$, $A\in \mathbb{R}^{m\times n}$,$X\in \mathbb{R}^{n\times p}$, $N_1$, $\epsilon$, $N_2$. 
   \STATE {\bfseries Output:} $A^{(+)}\in \mathbb{R}^{m\times n}$, and $X^{(+)}\in \mathbb{R}^{n\times p}$. 
   \STATE Rearrange column of $A$ and rows of $X$ such that such that $k^1\geq k^2\geq \dots \geq k^n$. %Sort $k_i$ such that $k_1\geq k_2\geq \dots \geq k_n$, and Set $X=0$.
  % \STATE Randomly pick the support $\Omega_i$, and the amplitude of $x^i$ such that $|\Omega_i|=k_i$. Randomly generate $A$.
    \REPEAT
	 \FOR{$i=1$ {\bfseries to} $N_1$}
		 \STATE Run Algorithm \ref{proc:innerRow} with input $\tilde{Y}=Y-[A\setminus a_i][X\setminus x^j]$, $x^i$, $a_i$, and $N$. Update $x^i$, and $a_i$.
	 \ENDFOR
   \STATE Rescale $A$ and $X$, such that $\forall i, \|A_i\|_2=1$.
	 \FOR{{\bfseries all $(_2^n)$ pairs of rows} \{i,j\}}
		 \STATE Run Algorithm \ref{proc:interRow} with input $\tilde{Y}=Y-[A\setminus \{a_i,a_j\}][X\setminus \{x^i,x^j\}]$, $x^i,x^j$, $a_i,a_j$. Update $x^i$ and $x^j$.
	 \ENDFOR
	 \STATE Run Algorithm \ref{proc:glAdj} with input $Y$, $A$, $X$, and $N_2$. Update $A^{(+)}$ and $X^{(+)}$.
   \UNTIL{$\|Y-AX\|_2^2-\|Y-A^{(+)}X^{(+)}\|_2^2\leq \epsilon$}
\end{algorithmic}
\end{algorithm}

%\subsection{Convergence of the Objective}

%\begin{theorem} The objective (\ref{alg:obj}) decreases monotonically and converges.
%\end{theorem}
%\subsection{Monotone Property of the Inner-Row Sparsity Adjustment}

\begin{lemma} The objective (\ref{alg:obj}) decreases monotonically in Algorithm \ref{proc:innerRow}.
\end{lemma}\label{lem:obj_dec}
\begin{proof}
Let 
%\begin{align}
$L(a_i,x^i,\Omega^i)= \|\tilde{Y}_i-a_ix^i(\Omega^i)\|_f^2$,
%\end{align}
where $\tilde{Y}_i=Y-\sum_{j\neq i}a_j x^j$. For monotonicity, it suffices to show $L(a_i^{(+)},x^{i(+)},\Omega^{i(+)})\leq L(a_i,x^i,\Omega^i)$.

Since
%\begin{align}
$L(a_i,x^i,\Omega^i) = \|\tilde{Y}_i-a_ix^i(\Omega^i)\|_f^2 =\|\tilde{Y}_{\Omega^i}-a_ix^i_{\Omega^i}\|_f^2+\|\tilde{Y}_{\Omega^{i\mathsf{c}}}\|_f^2$,%\end{align}
the second term $\|\tilde{Y}_{\Omega^{i\mathsf{c}}}\|_f^2$ is fixed given $\Omega^i$. Since $\{a_i^{(+)},x^{i(+)}\}$ minimizes $\|\tilde{Y}_{\Omega^i}-a_ix^i_{\Omega^i}\|_f^2$ for any given $\Omega^i$, we have
%\begin{align}
$L(a_i^{(+)},x^{i(+)},\Omega^i)\leq L(a_i,x^i,\Omega^i)$.
%\end{align}

In the second step $a_i^{(+)}$ is fixed, and w.o.l.g. let us assume $\|a_i^{(+)}\|_2=1$. We would like to
\begin{align}
&\min_{\Omega^i,x^i} \|\tilde{Y}-a_i^{(+)}x^i\|_f^2
=\min_{\Omega^i}\min_{x^i}\sum_{j\in {\Omega^i}} \|\tilde{y_j}-a_i^{(+)}x^i(j)\|_2^2+\sum_{j\notin {\Omega^i}} \|\tilde{y_j}\|_2^2\nonumber\\
=&\min_{\Omega^i}\sum_{j\in {\Omega^i}} \min_{x^i_{\Omega_i}}\|\tilde{y_j}-a_i^{(+)}x^i(j)\|_2^2+\sum_{j\notin {\Omega^i}} \|\tilde{y_j}\|_2^2%\nonumber\\
%=\min_{\Omega^i}\sum_{j\in {\Omega^i}} \|\tilde{y_j}-a_i^{(+)}a_i^{(+)T}\tilde{y_j}\|_2^2+\sum_{j\notin {\Omega^i}} \|\tilde{y_j}\|_2^2\nonumber\\
=\min_{\Omega^i}\sum_{j\in {\Omega^i}} (\|\tilde{y_j}\|_2^2-(a_i^{(+)T}\tilde{y_j})^2)+\sum_{j\notin {\Omega^i}} \|\tilde{y_j}\|_2^2\nonumber\\
=&\sum_j\|\tilde{y_j}\|_2^2-\max_{\Omega^i}\sum_{j\in {\Omega^i}} (a_i^{(+)T}\tilde{y_j})^2%\nonumber\\
=\|\tilde{Y}\|_f^2-\max_{\Omega^i} \|a_i^{(+)T}\tilde{Y}_{\Omega^i}\|_2^2
\end{align}

If we would like to choose $k^i$ non-zeros in the $i$th row of $X$, then the optimal way of minimizing the objective is to choose the ones with the largest $|a_i^T\tilde{y_j}|$. Thus $\Omega^i$ corresponding to the $k^i$ entries with the largest $|a_i^T\tilde{y_j}|$s minimizes the reconstruction error. And the corresponding $x^i$ is determined by the projection of $\tilde{Y}_{\Omega^j}$ onto $a_i^{(+)}$.
\end{proof}

In a similar way, we prove the objective (\ref{alg:obj}) decreases monotonically in Algorithm \ref{proc:interRow}. The proof is omitted here.

\textbf{Convergence of the Objective:} 
Since the objective values generated by the algorithm is a monotonically decreasing sequence of non-negative real numbers, we know it converges according to the monotone convergence theorem.

\section{Dictionary Initialization}\label{sec:init}

The proposed algorithm, though has a convergent objective, is still a local one. A natural question is how we should initialize the sparsity pattern of $X$.  
We use a simple batchwise iterative procedure to generate the initialization of the dictionary for Algorithm \ref{alg:monosvd}. 

The initialization procedure is listed in Algorithm (\ref{alg:dictApprox}), where OMP is Orthogonal Mathing Pursuit, and OMP ($\mathrm{vec}(Y), I_p\otimes A, N$) treats the block of samples as a whole compared to the sample-by-sample way in $K$-SVD. %As listed in Algorithm \ref{alg:dictApprox}, 
The input of the algorithm (\ref{alg:dictApprox}) includes the total number of non-zeros $N$ in the representation $X$ of a batch of samples $Y$. There is no guarantee that the Dictionary Approximation algorithm converges, but empirically it provides a good initialization for our Batch-SVD algorithm.

\begin{algorithm}[tb]
   \caption{Dictionary Approximation using Block OMP}
   \label{alg:dictApprox}
\begin{algorithmic}
   \STATE {\bfseries Input:} $Y\in \mathbb{R}^{m\times p}$, $A_o$, $N$, $T$. 
   \STATE {\bfseries Output:} $A\in \mathbb{R}^{m\times n}$ and $X\in \mathbb{R}^{n\times p}$. 
    \STATE Initialize $X=0$, and $A=A_o$.
   \FOR{$t=1$ {\bfseries to} $T$}
      \STATE (1) $X\leftarrow$ OMP ($\mathrm{vec}(Y), I_p\otimes A, N$).%Block OMP(Y, A, N).
      \STATE (2) $A\leftarrow \arg\min_A\|Y-AX\|_f^2$.
   \ENDFOR
\end{algorithmic}
\end{algorithm}

\section{Experiments}\label{sec:exp}
In this section, we compare our proposed approach to state-of-the-art dictionary learning algorithms on real world data sets, including natural image patches and general machine learning sets. The focus of the experiments is data compression. Data compression is a critical application for dictionary learning. Particularly in the big data regime, if the samples are represented using only a few coefficients, great storage space can be saved. It may also be used in signal communication, where the sender and receiver keep a copy of the dictionary, and only the sparse coefficients are transmitted. In the experiment, we choose a dictionary $A$ and a sparse coefficient matrix $X$, and try to minimize the reconstruction error $\|Y - A X\|_f$ with a given number of non-zeros in $X$. The number of non-zeros is set the same as that produced by $K$-SVD and online dictionary learning, and we compare the reconstruction errors.

\subsection{Data Preparation}
We use $10$ data sets in our experiments. The first one is the demonstration image set provided in the $K$-SVD toolbox \cite{Ron06}, with $5$ images: Barbara, boat, house, Lenna, and peppers. For each image we randomly sample $3000$ overlapping patches of size $8$-by-$8$ as the training set, and use another randomly sampled $3000$ patches as the testing samples in the open-set evaluation. The second data set is the Notre Dame Image library which contains $715$ images taken from the Notre Dame Cathedral in Paris. To make the scales consistent, we resize each image to $512$-by-$512$, and then randomly sample $10,000$ patches as the training set. In the testing stage, we randomly sample $3000$ image patches from the image library for a total of $100$ runs, and report the mean and standard deviation of the reconstruction error. %and use another randomly sampled $10,000$ patches in the open set evaluation. 
We also carry out experiments on $8$ UCI data sets, including mnist, iris, yeast, glass, wine, ecoli, liver-disorder, and heart-disease\footnote{We remove the sample columns with `nan' entries in the heart-disease data set.}.

\begin{figure} [tbp]
\centering
\includegraphics[width=6cm]{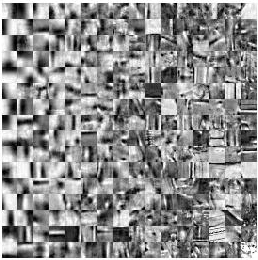}
\includegraphics[width=6cm]{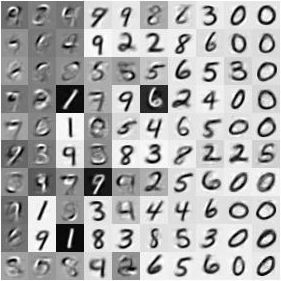}
\caption{Demonstration of dictionary atoms learned using our algorithm. The left is the learned dictionary from random patches in the Notre Dame library, and the right is from MNIST digit set.}\label{Fig:dictDemo}
\end{figure}

\subsection{Demonstration of the Dictionary}

For a better illustration we train a square dictionary using $10,000$ randomly sampled image patches of $16$-by-$16$ from the Notre Dame Image data set, and thus the dictionary $A$ has dimensions $256$-by-$256$. The average number of non-zeros per sample is $\|X\|_0/p\approx 2.0223$, and $n=256$. For the MNIST set, we randomly sample $3000$ images from the training set, and learn a $784$-by-$100$ dictionary. The average number of non-zeros per sample is around $6$.

\begin{table}
\centering
\caption{Reconstruction Errors on Natural Image Patches. The digit outside the bracket is the average $L_2$ norm of the errors per patch, and the digit inside the bracket is the standard deviation.}\label{tbl:img_close}
%\centering
\subtable[Comparison with $K$-SVD]{
       \begin{tabular}{|l|c||c|c|c|c|}
       \hline
$\times 10$&\textbf{online}&\textbf{KSVD}&\textbf{Wvlet}&\textbf{Rnd}&\textbf{Batch}\\\hline
\textbf{barbara}&2.9(0.8)&2.0(1.0)&3.1(2.5)&7.4(5.7)&1.8(0.4)\\\hline
\textbf{boat}&3.2(0.6)&2.1(0.6)&3.0(1.9)&6.4(5.6)&1.9(0.3)\\\hline
\textbf{house}&2.2(0.8)&1.7(0.9)&3.1(2.7)&6.8(8.3)&1.5(0.4)\\\hline
\textbf{lena}&2.5(0.6)&1.8(0.7)&2.7(2.0)&6.1(6.2)&1.7(0.3)\\\hline
\textbf{peppers}&2.9(0.6)&2.1(0.7)&3.0(1.8)&6.2(6.7)&2.0(0.3)\\\hline
\textbf{ND}&4.3(2.4)&3.1(2.6)&4.0(3.8)&8.2(7.9)&2.7(1.4)\\\hline
%\hline
       \end{tabular}
       \label{tab:firsttable}
}
%\qquad
\subtable[Comparison with Error-based KSVD]{        
       \begin{tabular}{|c|c|c|c|c|}
\hline
\textbf{ESVD}&\textbf{KSVD}&\textbf{Wvlet}&\textbf{Rnd}&\textbf{Batch}\\\hline
2.5(0.4)&3.1(1.7)&5.0(4.1)&10.3(8.1)&2.4(0.6)\\\hline
2.7(0.3)&3.2(1.1)&4.7(3.3)&8.8(7.8)&2.6(0.4)\\\hline
2.2(0.8)&2.7(1.9)&5.0(5.1)&8.6(10.4)&2.1(0.8)\\\hline
2.4(0.4)&3.0(1.8)&5.0(4.9)&8.3(8.7)&2.3(0.6)\\\hline
2.6(0.3)&2.9(1.1)&4.3(3.4)&7.9(8.6)&2.5(0.5)\\\hline
2.4(0.8)&2.7(2.2)&3.4(3.2)&7.2(6.6)&2.1(1.0)\\\hline       
\end{tabular}
       \label{tab:secondtable}
}
\end{table}

\subsection{Reconstruction on Natural Image Patches}

We compare our algorithm with the online dictionary learning, $K$-SVD, and the overcomplete wavelets with orthogonal matching pursuit. Also the result using a Gaussian random dictionary with OMP is presented as a baseline. 
For the online dictionary learning, we set $\lambda=10$. \footnote{We use SPAMS \cite{SPAMS} with the default batch size 512 in our evaluation.} Since the sparsity for the online dictionary learning is only softly constrained, we first run the online dictionary learning algorithm and then force $k=\lfloor\|X_{online}\|_0/p\rfloor$ in the $K$-SVD algorithm, such that the total number of non-zeros in the representation derived using $K$-SVD is not larger than that of the online learning algorithm. We then set the number of non-zeros in our algorithm to be exactly the same as the $K$-SVD algorithm. %When $\lambda=10$, the average non-zeros per sample is $k=9$. 
The iteration number of online learning and $K$-SVD is set as $100$. The iteration number for our algorithm is set at $20$ with $N_1=3$ and $N_2=10$. To accelarate the algorithm, the inter-row support switching is only carried out when the objective decrement in the inner-row adjusment is smaller than $0.05$. The iteration number of the initialization precedure (\ref{alg:dictApprox}) is set at $80$. For the image data sets, the number of atoms in the dictionary $A$ is $n=256$.

\begin{table}
\caption{Reconstruction Errors on the UCI Data Sets. The digit outside the bracket is the average $L_2$ norm of the errors per sample, and the digit inside the bracket is the standard deviation.}\label{tbl:UCI_close}
\centering
\subtable[Comparison with $K$-SVD]{
       \begin{tabular}{|l|c||c|c|c|}
\hline
$\times 10^{-3}$&\textbf{online}&\textbf{KSVD}&\textbf{Rnd}&\textbf{Batch}\\\hline
\textbf{liver}&22.6(4.0)&7.2(6.9)&51.9(35.6)&5.8(7.3)\\\hline
\textbf{iris}&20.1(0.1)&522.1(275.9)&102.1(46.3)&11.1(12.4)\\\hline
\textbf{yeast}&27.8(7.5)&8.1(8.6)&36.2(25.6)&7.9(8.5)\\\hline
\textbf{glass}&20.1(0.2)&57.6(26.7)&68.5(35.4)&0.9(0.5)\\\hline
\textbf{wine}&20.1(0.1)&64.8(30.9)&258.1(10.5)&1.6(0.7)\\\hline
\textbf{ecoli}&27.7(3.5)&8.7(9.5)&50.4(37.3)&2.9(3.7)\\\hline
\textbf{heart}&21.3(0.9)&173.6(103.5)&345.6(33.8)&8.2(2.7)\\\hline
       \end{tabular}
       \label{tab:firsttable-2}
}
%\qquad
\subtable[Comparison with Error-based KSVD]{        
       \begin{tabular}{|c|c|c|c|c|}
       \hline
\textbf{ESVD}&\textbf{KSVD}&\textbf{Rnd}&\textbf{Batch}\\\hline
0.8(1.9)&7.4(6.5)&40.7(24.3)&0.6(1.3)\\\hline
491.0(309.7)&331.4(287.1)&152.1(55.4)&2.6(4.1)\\\hline
2.0(2.9)&14.5(15.0)&37.2(27.4)&3.1(4.9)\\\hline
50.7(65.4)&45.7(24.3)&442.3(11.7)&3.7(2.3)\\\hline
80.0(45.7)&66.4(39.6)&797.2(1.5)&4.8(2.8)\\\hline
1.6(2.4)&16.7(17.9)&41.0(28.7)&2.5(3.5)\\\hline
172.7(82.4)&182.7(86.0)&334.4(22.6)&5.1(1.6)\\\hline    
\end{tabular}
       \label{tab:secondtable-2}
}
\end{table}

\subsection{Reconstruction on UCI Data Sets}
We also carried out experiments on the UCI data sets. For all the algorithms, we set the number of atoms in the dictionary $n=30$. The data vectors are normalized to have unit norm before feeding into the algorithms. Again we first run the online dictionary learning algorithm with $\lambda=0.02$, and then set $k=\lfloor \|X_{online}\|_0/p\rfloor$, where $X_{online}$ is the coefficient derived using the online learning algorithm. We set the same number of non-zeros for our batch dictionary learning algorithm as that produced by $K$-SVD. %The average number of non-zeros for the $7$ UCI sets are $4$,$2$,$6$,$5$,$4$,$5$,$4$
The reconstruction errors are listed in the first part of Table (\ref{tbl:img_close}) and Table (\ref{tbl:UCI_close}). We can see that the batchwise algorithm works consistently better than the other methods.
\subsection{Reconstruction-Error Based $K$-SVD}

We also compared with the reconstruction-error-based $K$-SVD (ESVD) algorithm proposed in \cite{Rubinstein08}. Since it is not easy to control exactly the number of non-zeros produced by ESVD, again we we first run ESVD and then set the number of non-zeros in our algorithm to be exactly the same as the ESVD algorithm. For comparison the reconstruction errors of the original $K$-SVD, wavelets, and random dictionary are also presented with $k=\lfloor \|X_{ESVD}\|_0/p\rfloor$. For the Notre Dame library we set the reconstruction error $\epsilon=30$ , yielding an average sparsity $k\approx 9$ per sample. For the UCI data sets we set $\epsilon=0.01$. The results are presented in the second part of Table (\ref{tbl:img_close}) and (\ref{tbl:UCI_close}), from which we observe that the ESVD algorithm performs reasonably better than the original $K$-SVD algorithm. The batchwise algorithm, with the same sparsity level, gives better approximations on all the sets except yeast and ecoli.

\section{Conclusion}\label{sec:conclu}

In this paper we propose a monotone dictionary learning algorithm that is optimized for sample batches. The reconstruction error is minimized by a series of support switching procedures withing the sample batch. We prove the objective monotonically decreases and converges in the support switching procedures. Using the proposed block orthogonal matching pursuit algorithm as a warm start, the batchSVD algorithm gives a better approximation in terms of the reconstruction error at the same level of sparsity. 

\bibliographystyle{plain}
\bibliography{batchsvd}

\end{document}